\documentclass[conference]{IEEEtran}
\IEEEoverridecommandlockouts
\usepackage{cite}
\usepackage{amsmath,amssymb,amsfonts}
\usepackage{algorithmic}
\usepackage{graphicx}
\usepackage{textcomp}
\usepackage{xcolor}
\usepackage[inline]{enumitem}
\usepackage{float}
\usepackage{booktabs}
\usepackage{multirow}
\newcommand{\ra}[1]{\renewcommand{\arraystretch}{#1}}
\def\BibTeX{{\rm B\kern-.05em{\sc i\kern-.025em b}\kern-.08em
    T\kern-.1667em\lower.7ex\hbox{E}\kern-.125emX}}
\begin{document}

% Tentative Title
% \title{Machine Learning to Identify Key Value Drivers in Energy Markets}
%\title{A Machine Learning Framework to Deconstruct the Primary Drivers for Price Events in Electricity Markets with High VRE
\title{A Machine Learning Framework to Deconstruct the Primary Drivers for Electricity Market Price Events
\thanks{Pacific Northwest National Laboratory is operated for DOE by Battelle Memorial Institute under Contract DE-AC05-76RL01830. This work was supported by DOE's Water Power Technology's Office. } 
\thanks{Corresponding author e-mail: milan.jain@pnnl.gov.}
\\[-1.0ex]
}
% \IEEEaftertitletext{\vspace{-5\baselineskip}}
\author{
\IEEEauthorblockN{Milan Jain, Xueqing Sun, Sohom Datta and Abhishek Somani}
\IEEEauthorblockA{Pacific Northwest National Laboratory, Richland, WA, USA}
% \IEEEauthorblockA{\{milan.jain, xueqing.sun, sohom.datta, abhishek.somani\}@pnnl.gov}
\\[-100.0ex]
}

\maketitle

\begin{abstract}

Power grids are moving towards 100\% renewable energy source bulk power grids, and the overall dynamics of power system operations and electricity markets are changing. The electricity markets are not only dispatching resources economically but also taking into account various controllable actions like renewable curtailment, transmission congestion mitigation, and energy storage optimization to ensure grid reliability. As a result, price formations in electricity markets have become quite complex. Traditional root cause analysis and statistical approaches are rendered inapplicable to analyze and infer the main drivers behind price formation in the modern grid and markets with variable renewable energy (VRE). In this paper, we propose a machine learning-based analysis framework to deconstruct the primary drivers for price spike events in modern electricity markets with high renewable energy. The outcomes can be utilized for various critical aspects of market design, renewable dispatch and curtailment, operations, and cyber-security applications. The framework can be applied to any ISO or market data; however, in this paper, it is applied to open-source publicly available datasets from California Independent System Operator (CAISO) and ISO New England (ISO-NE).

%These changes in operations paradigm lead to increased volatility in electricity market prices. For utilities and other market participants to remain competitive in their bidding strategy, it is important to understand the root causes of various price spikes and price volatility. Market participants often have to rely on the publicly available data set for their analysis,  and these publicly available data sets may not be sufficient in understanding the root causes of price behaviors. In this paper, a robust methodology is developed to identify the various system conditions that lead to extreme prices in the electricity market using the help of data analysis and machine learning techniques. The methodology can be extended to any electricity market scenario. The overall results from this methodology using realistic data set from the California Independent System Operator (CAISO) publicly available data set suggest that renewable generation and forecast errors have significant effect on the economics of the electricity market. The correlations between electricity market prices and renewable generation will increase further as the industry move towards a net zero carbon and net zero marginal cost electricity market.

\end{abstract}


% \section{Related Work}
% The related work goes here. 

% \section{CAISO Data}
% Describe data here. 

\section{Introduction}
With the increasing penetration of renewable energy resources, operating an electric grid is becoming complex~\cite{jain2015combining}. It often leads to differences in planned versus actual operations between day-ahead and real-time markets. Various market instruments like virtual bidding, reserve markets, flexible ramping products, and demand response provide mechanisms to achieve convergence between the markets and manage imbalances and uncertainties between day-ahead and real-time markets \cite{caiso_price}. However, even with the multitude of these advanced market instruments in place, one or more system operating conditions occurring in different temporal and spatial combinations can still cause price volatility and price spikes. 

\subsection{Related Work}
While there exists a significant amount of work on price formation, electricity price forecasting, and price spike detection, only a handful of studies have explored the interpretation of price spike events. In most cases, studies have mainly focused on understanding price formation in specific energy markets, which cannot be generalized to other markets. For instance, Goncalves et al. \cite{8810763} applied a set of explanatory models in the MIBEL electricity market spot price to understand the main drivers of electricity price. Velasco et al.~\cite{6972411} proposed a graphical analysis to visualize key variables associated with the European electricity market prices. In contrast, our proposed framework is generic and can be applied to any energy market. For any generic market, the framework proposed by Gonzalez et al.\cite{gonzalez2015important} is also useful, where the feature importance of trained models is studied to improve the accuracy of price forecasting. 
% However, these studies specifically focused on the statistical analysis of price formation in the European markets with less focus on anomalous price events (such as price spikes). Moreover the key takeaways cannot be generalized across ISOs and other energy markets - an important lesson that we also noticed in our study when comparing CAISO and ISO-NE. 
% Closest to our work is the study by Gonzalez et al.\cite{gonzalez2015important}, where feature importance of trained random forest models were studied to improve the accuracy of price forecasting. 
Our proposed framework 
% while inheriting key components of some of the existing studies to identify the system state (analyzing feature importance from \cite{gonzalez2015important} and clustering from \cite{mori2007data}), 
is an extension of those studies and incorporates advanced concepts of interpretable machine learning to automatically deconstruct the primary drivers of price spike events.  
%, who currently analyze such events after the fact since they have to perform price corrections and adjustments during settlement process, which is not possible in real-time. % The framework is applied to two energy markets in USA but can be extended to any other energy market with minimal effort.
% Since the framework is data-These insights and key drivers for price spikes can be used by market and system operators to analyze market conditions in real-time. Often, market operators have to analyze market data after the fact and have to perform various price corrections and adjustments during settlement process as analyzing complex market data in real-time in not trivial. This framework allows market operators can cluster every market run results into clusters and validate the market results close to real-time. The cluster analyses results can be utilized to label price-spike events automatically and assign a specific reasons like congestion, renewable volatility, forecast errors, etc. as primary drivers. This framework can be further utilized to eliminate cyber-security related market attacks which can result in price spikes as well as  by market designers and policy analysts in understanding the market mechanisms for price formation, improving the forecast accuracy of electricity prices, bringing transparency in the market operations, and designing appropriate market-based interventions to mitigate such scenarios. 
% \vspace{-0.2cm}
\subsection{Challenges}
The challenges in designing such a framework, specifically in the context of price spikes, include: (a) rare occurrences of price spike events; 
(b) complex interactions between multiple system conditions leading to price spikes; and (c) limited data availability due to the confidential nature of price bids and generator availability. 

\subsection{Proposed Framework}
In the context of this study, an event implies a price spike event, and a price spike is defined as a segment of time in which the marginal cost of energy exceeds a certain threshold. The proposed comprehensive analytical framework integrates price-spike detection, statistical analysis of detected events, feature engineering, training of machine learning (ML) model, and automatic identification of key drivers driving the price-spike using the trained model. The framework also allows users to understand the system state during those events by clustering the energy market data using an unsupervised ML algorithm and plotting them on a radar-chart visualization. The framework is applied to publicly available datasets collected from California ISO (CAISO) and ISO New England (ISO-NE). The insights related to their price spike events and the key drivers driving those events are discussed. 

% In this paper, we first develop a clear definition of a price spike event, which is a segment of time where electricity prices exceed a certain threshold. Then we propose a comprehensive analytical framework that integrates the detection of price-spike events, statistical analysis of these events, feature extraction of system state variables, and machine learning model-based identification of key drivers for the price-spike events. To allow user to understand the system state during those events, the framework clusters the data using unsupervised machine learning algorithms and provide radar-chart visualization of those states. The framework is applied to publicly available California ISO (CAISO) and ISO New England (ISO-NE) datasets and insights related to their price spike events and key drivers driving those events are discussed. 

The major contributions of this paper are:
1) a comprehensive machine learning-based framework for the detection and interpretation of price spikes, and
% which can be extended to study many other aspects such as curtailments in the power system and,
2) demonstrate the effectiveness of the proposed framework using data from two energy markets: CAISO and ISO-NE.
The proposed framework can be used by (1) market operators to analyze market conditions in real-time, (2) cyber-security experts to identify malicious spikes, and (3) market regulators to bring transparency into market operations.
\section{CAISO and ISO-NE Data Description}
%[mainly focus on the data itself]
% The proposed autoencoder-based root cause analysis framework for electricity prices is tested using 
% For this study, publicly available data from CAISO and ISO-NE were used to demonstrate the efficacy of the proposed framework. 
% though the framework can be extended to any electricity market without loss of generality 
% This section provides an overview of both datasets. 
% \begin{figure}[ht!]
%     \centering
%     \includegraphics[width=0.7\columnwidth]{figures/CAISO_ISONE_RESOURCEMix.png}
%     \caption{Resources Mix for CAISO and ISO-NE (2021)}
%     \label{fig:resourceMix}
% \end{figure}

% \vspace{-3mm}

% \vspace{-3mm}
\subsection{California ISO (CAISO)}
%CAISO manages a day-ahead market and an intra-day real-time market that economically dispatches generating resources to serve the forecast load, while managing various transmission and generation constraints.
CAISO is a \emph{nodal} market, which generates locational marginal prices (LMP) for over 4000+ price nodes throughout its footprint. In this paper, price spikes occurring at four major locational aggregate price (LAP) nodes for PG\&E (Pacific Gas and Electric Company), SCE (Southern California Edison), SDGE (San Diego Gas and Electric), and VEA (Valley Electric Association) are analyzed. The market and operational data for CAISO for the years  2018, 2019, and 2020 were utilized to apply the framework.

\subsection{ISO New England (ISO-NE)}
% nearly 25 years, New England’s wholesale electricity markets have attracted billions of dollars in private investment in some of the most efficient, lowest-emitting power resources in the country—providing reliable electricity every second of every day,lowering wholesale prices. 
For ISO-NE, natural-gas-fired generation, nuclear, other low- or no-emission sources, and imported electricity (mostly hydropower from Eastern Canada) provided the region’s electricity in 2021.%, as seen in Fig.~\ref{fig:resourceMix}. 
The total generation by renewable energy is 12.44\%, including 4\% by wind and 3\% by solar.
% 
% \subsubsection{Data Description}
We collected market and operational data from ISO-NE for 2020 and 2021. The LMPs and corresponding energy, congestion, and loss components of LMPs are obtained from ISO-NE for the eight load zones and hubs. 

\begin{figure}[ht!]
    \centering
    \includegraphics[width=0.6\columnwidth, height=3.5cm]{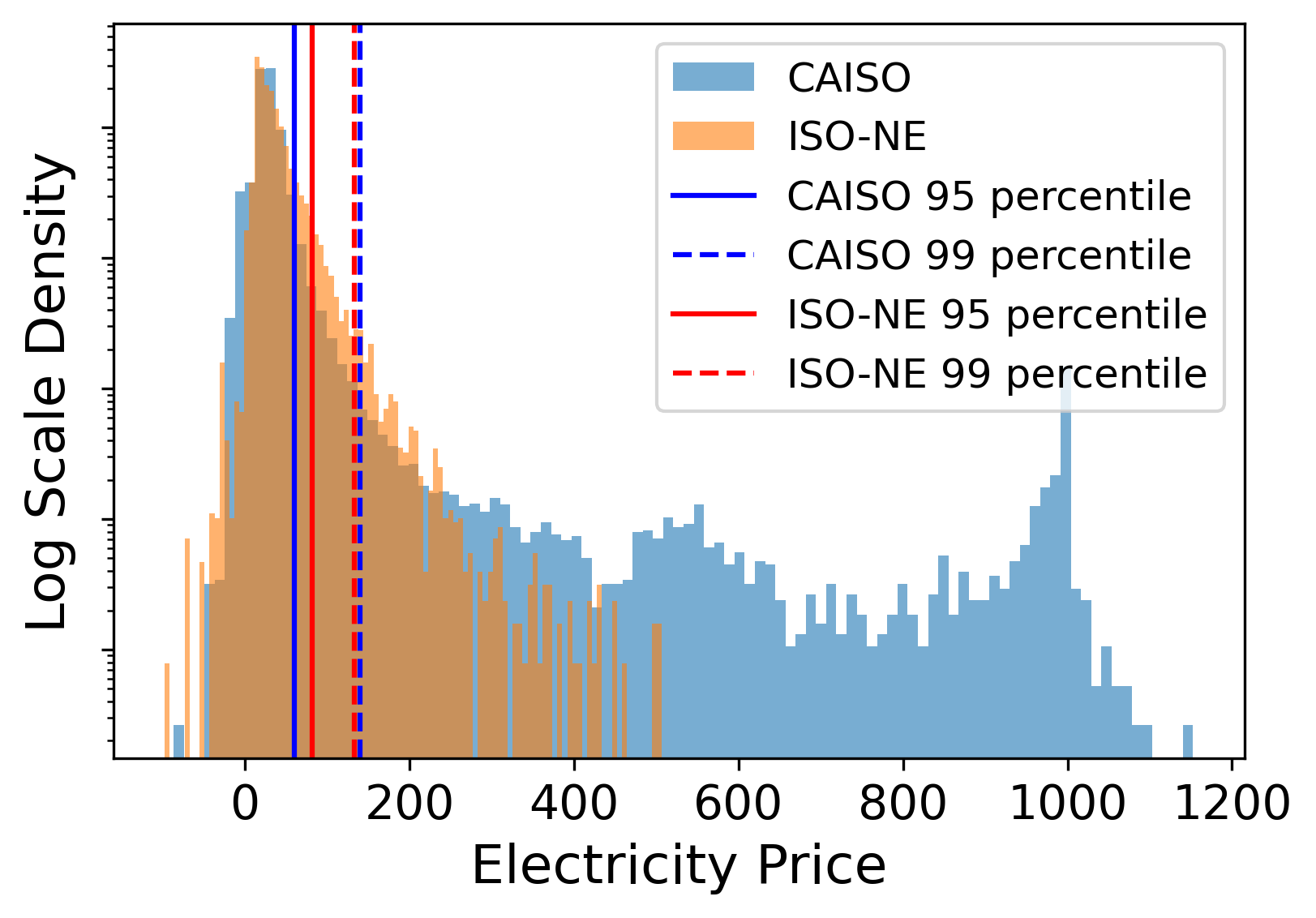}
    \vspace{-0.3cm}
    \caption{Plot and Histogram of Energy Component of LMP in CAISO}
    \label{fig:ec_dist_plot}
    \vspace{-0.4cm}
\end{figure}

In this analysis, we only focus on identifying the key drivers that impact the marginal cost of energy. Fig.~\ref{fig:ec_dist_plot} compares the distribution of the energy component of LMP between CAISO and ISO-NE. Compared to CAISO, energy prices in ISO-NE are relatively lower, and the price spikes are sparser. %The price spikes happen most in the summer months, and some in winter. Compared with 2020, 2021 has experienced more price spikes, shown in Fig.~\ref{fig:isone_ec_bymonth}.

% \begin{figure}[ht!]
%     \centering
%     \includegraphics[width=0.8\columnwidth]{figures/isone_plot_hist_energycomponent.png}
%     \caption{Plot and Histogram of Energy Component of LMP in ISO-NE}
%     \label{fig:isone_ec_plot}
% \end{figure}
\section{Methodology}
\begin{figure}[ht!]
    \centering
    \includegraphics[width=0.8\columnwidth]{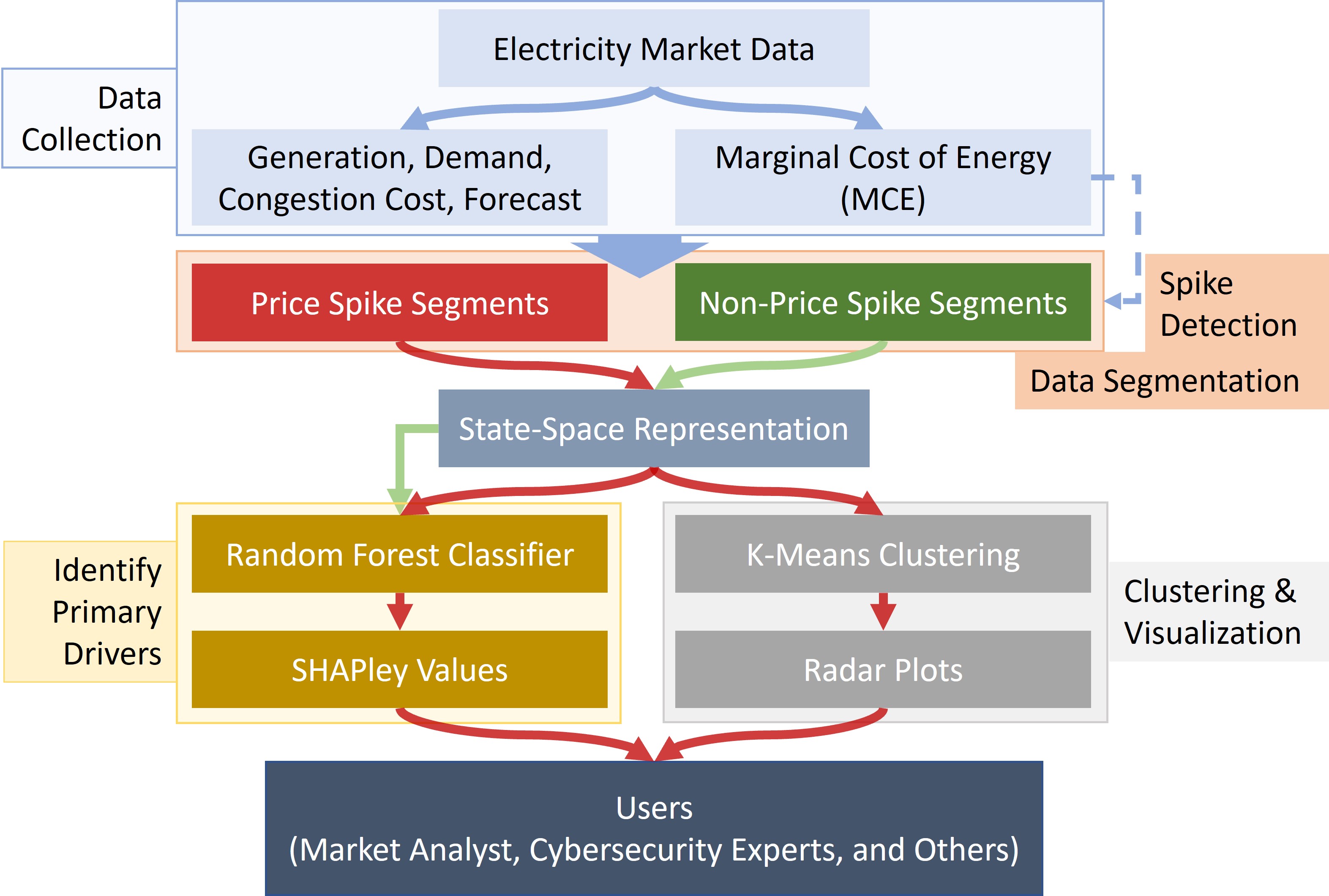}
    \vspace{-0.2cm}
    \caption{System architecture}
    \label{fig:rca_flowchart}
    \vspace{-0.6cm}
\end{figure}

% To detect the electricity price spikes and analyze the underlying key drivers, we develop a comprehensive 
The proposed framework 
(as shown in Fig.~\ref{fig:rca_flowchart}) 
begins with the statistical analysis of the market data for price spike detection and divides the data into price spike and non-price spike segments. Next, those segments are summarized into a state-space representation, which is used to train an ML model to classify spike segments from non-spike segments. 
% identify key features highly correlated with the price spike segments. 
The predictions generated by the trained classifier are next analyzed using SHAP (SHapley Additive exPlanations)~\cite{lundberg2017unified} - a game theoretic approach used to 
% generate posh-hoc explanations for the output of any machine learning model. These explanations (aka Shapley values) 
quantify the impact (both positive and negative) of each feature on the model outcome.
% - whether a particular data segment is spike/non-spike, in the context of this study study.
Finally, to provide user context about the state of the energy market, an unsupervised clustering algorithm is used to cluster the market data and define the system state, which is visualized using radar charts. 

% Next, the framework is discussed in detail. 
% First we depict the overall framework, then we describe the key elements of the framework in detail in the subsequent subsections.
% \begin{itemize}
%     \item overall framework or flowchart
%     \item price spike detection/segmentation
%     \item State space representation of the variables (why we do it and the benefits of it)
%     \item root cause analysis with focus on autoencoder and clustering
% \end{itemize}
% \subsection{Deep Learning Based Root Cause Analysis Framework}
% As shown in \ref{fig:rca_flowchat}, our proposed framework integrates [] 
% \vspace{-2cm}

% \begin{figure}[ht!]
%     \centering
%     \includegraphics[width=1\columnwidth]{figures/data_segmentation_for_price_pikes.pdf}
%     \caption{llustration of Data Segmentation: CAISO}
%     \label{fig:dataSeg}
% \end{figure}
\subsection{Spike Detection and Data Segmentation}
A \emph{price-spike point} is a price in time that exceed a certain threshold. The threshold can be a fixed number, or can be computed from the data using percentiles.
% such as a price at the 95$^{th}$ percentile. 
Once detected, the price-spike points closed to each other are grouped together to define a \emph{price-spike event}. Next, the data is divided into segments: normal data segments (no price-spike event) and anomalous data segments (atleast one price-spike event). For the anomalous segments, data between $[t_{first}-b_{len},t_{last}+f_{len}]$ is selected, where $t_{first}$ and $t_{last}$ depict the first and last occurrence of spike in the grouped event, and $b_{len}$ and $f_{len}$ captures the recent history and near future around the spike. 
% These segments help us analyze the cause and the effects of spikes.
Rest of the data is divided into hourly normal data segments.
% , as shown in Fig.~\ref{fig:dataSeg}. 

% Time series price data are segmented to price spike segments and “regular” price segments. Figure \ref{fig:dataSeg} shows the concept of data segmentation and the steps are as below:
% \begin{itemize}
%     \item In this step, we first group the events which are close to each other. 
%     \item In the next step, we fetch the anomalous data segments. In the anomalous data segments, we select data between $[t_{first}-b_{len},t_{last}+f_{len}]$. Here, $t_{first}$ and $t_{last}$ depict the first and last occurrence of spike in the grouped event, and $b_{len}$ and $f_{len}$ show the backward and forward size of the data. These segments help us analyze the cause and the effects of spikes.
%     \item In the third step, we divide the rest of the data in normal data segments. 
% \end{itemize}

\subsection{State Space Representation}
% Therefore, in addition to analyzing raw time-series, we analyzed the evolution of system state computing a few summary statistics shown in Table~\ref{table:state_space}. For every time-series, we computed the summary statistics for a fixed time-period just before the price spike. We define this time period as \emph{segment}. 
Once the data is divided into segments, instead of using raw signals, derived features are computed to capture different aspects of the features. The derived features include 
\begin{enumerate*}
    \item \emph{mean:} to capture average value of a feature within the segment,
    \item \emph{std:} to capture feature volatility within the segment,
    \item \emph{average gradient:} to quantify trend in the feature value,
    \item \emph{maximum gradient:} to measure sudden change in the feature value within the segment. 
\end{enumerate*}
The sign of average and maximum gradient indicates whether the change was positive or negative.

\subsection{Identifying Primary Drivers}
The state-space representation is next used to train a Random Forest Classifier
% ~\cite{breiman2001random} 
to classify spike/non-spike segments. Random Forest is an ensemble learning method that constructs multiple decision trees and each tree captures simple decision rules inferred from the data features. 
% Random forest is robust to classical problems (such as multicollinearity) since prediction is based on voting by the decision trees. 
Though the learned rules (across the trees) can be used to highlight most important features (as done by \cite{gonzalez2015important}), the information cannot be used to quantify the feature importance for individual predictions. 

\begin{figure}[ht!]
    \centering
    \includegraphics[width=\columnwidth, trim={2cm 4.75cm 0 0},clip]{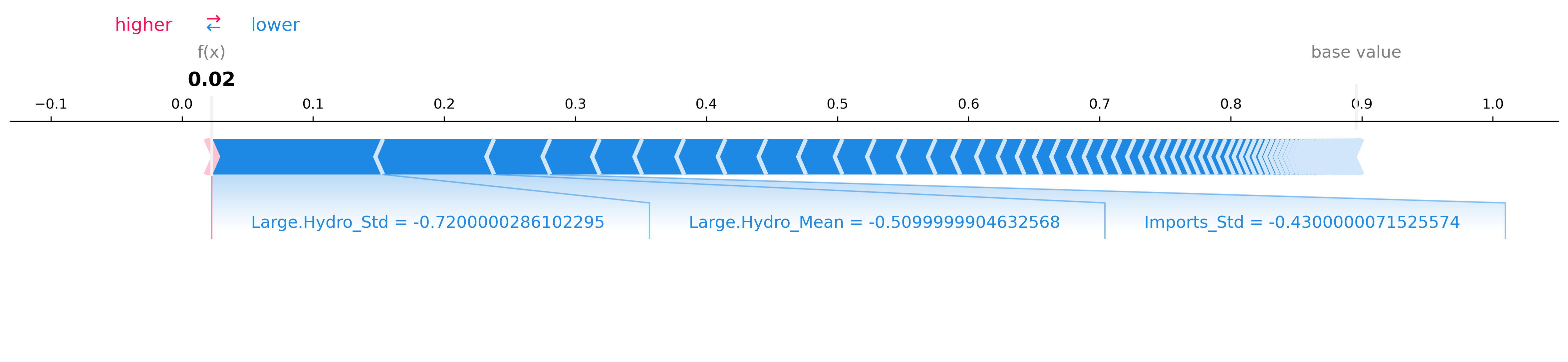}
    \includegraphics[width=\columnwidth, trim={5cm 4.75cm 0 0},clip]{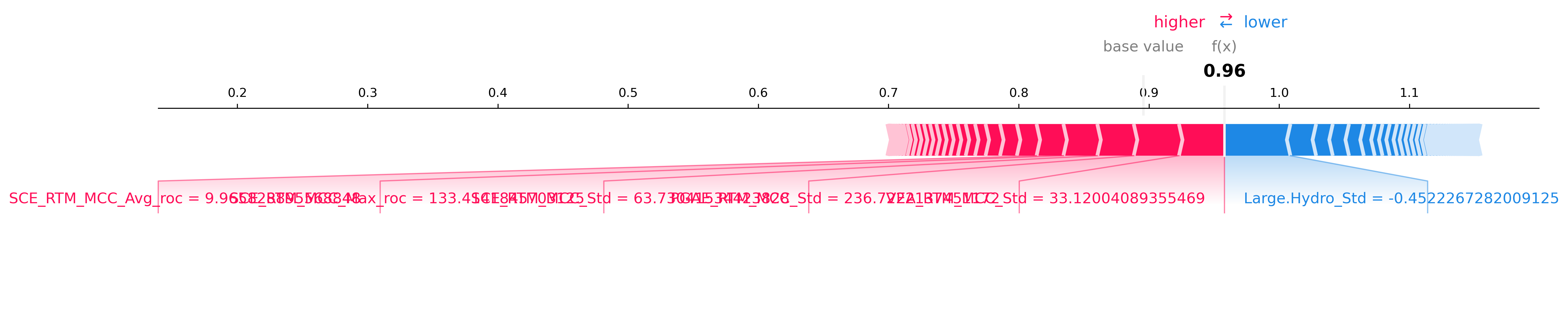}
    \caption{Feature importance through Shapley values}
    % : [top] a normal segment where features in blue are pushing prediction towards 0 (indicates normal segment) [bottom] an anomalous price-spike segment where features in red are pushing prediction towards 1 (indicates price-spike segment).}
    \label{fig:shap_explain}
    \vspace{-0.5cm}
\end{figure}

In this study, we use SHAP (SHapley Additive exPlanations)~\cite{lundberg2017unified} to determine the most important features and their influence on the model prediction. For any segment (normal/anomalous), Shapley values quantify the influence of individual features on the prediction generated by the trained model. For instance, Fig.~\ref{fig:shap_explain} compares shapley values of features between a normal (top) and an anomalous (bottom) segment from CAISO. While features in blue are pushing the prediction towards being normal, features in red are driving prediction towards being anomalous. It is evident from the plot that Shapley values of features advocating for spike is almost non-existent for the normal segment (top), however, significantly higher for the anomalous segment (bottom), and vice-a-versa for Shapley values of features in blue advocating for segment being normal.

To identify key drivers, framework uses the top five features from the red category, sorted by their contributions towards the predicted value and present it to the user.

\subsection{Clustering and Visualization}
Most often, the complex interaction between multiple components of an energy market lead to anomalous price behaviors. While it is important to identify key variables associated with a price spike event, it is hard to make sense out of that information without knowing the context - the system state. Therefore, the framework uses K-Means clustering on the state-state representation of the data to identify potential system states and visualize the output using radar plot. This additional feature provides some context to the user about the complex interplay between different components of the market and help them take an informed decision. 

The framework and the data will be open-sourced as part of this study. The links have been anonymized for the review. 
% The radar plot with information about the system state provides context to the user when analyzing the key features related to a price-spike event.

% find correlations between price spike segments and any significant deviation in other signal values. In the current framework, we implemented one supervised machine learning model – Random Forest Classifier and one unsupervised deep learning model – Autoencoders to identify the primary drivers related to price spike events. 
% In supervised learning, a random forest classifier is trained to classify spike/non-spike segments and the estimations are explained through SHAP values\cite{} to identify primary drivers for price spike events. On the other hand, autoencoders (AEs) are a special type of artificial neural network (also referred to as ANN) with one or more hidden layers to learn a latent representation of the input data. The latent representation is used to reconstruct the input. 
% Autoencoder for anomaly detection is an unsupervised ML technique, the input data does not need to be labelled and little prior knowledge about input data is necessary. 
% When trained on normal data instances, autoencoders fail to reconstruct abnormal inputs, and therefore a higher reconstruction error (RE) is observed for anomalous signals. 
% [Autoencoder-based anomaly root cause analysis for wind turbines])
% In the next Section, the application of the algorithms and its evaluation are presented.

\section{Evaluation}
% The section title: Root cause analysis on CAISO and ISO-NE results
% \begin{itemize}
%     \item 
% \end{itemize}
\begin{figure*}[ht!]
    \centering    
    \includegraphics[width=0.85\textwidth]{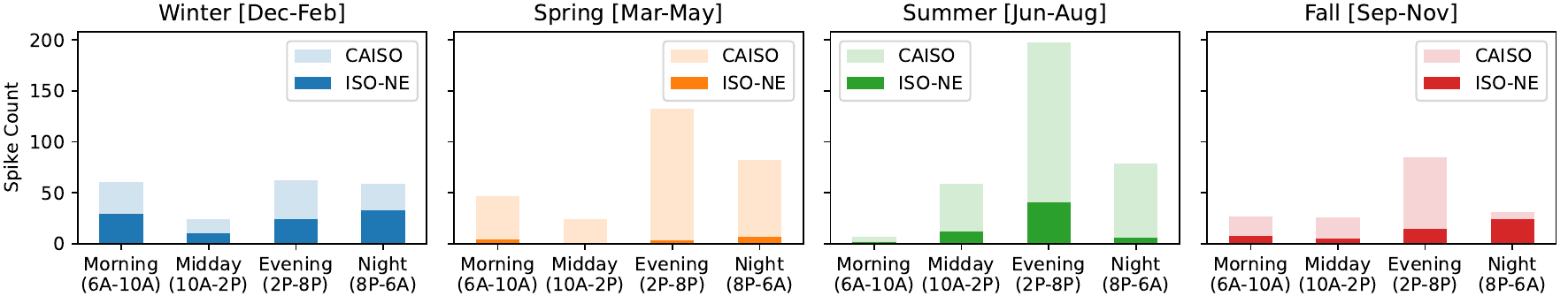}
    \includegraphics[width=0.65\textwidth]{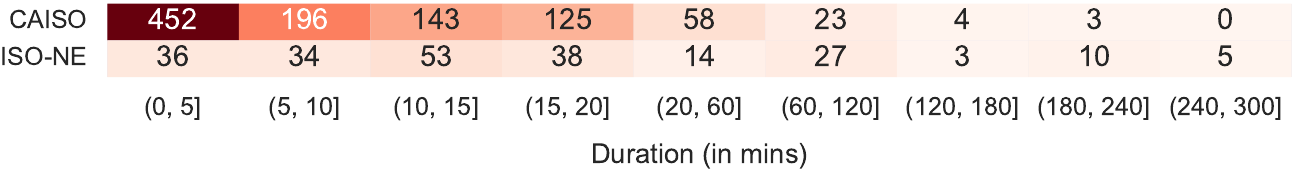}
    \vspace{-0.2cm}
    \caption{Spike Distribution: [top] Distribution of spike events across seasons and different times of the day. [bottom] Distribution of spike events by duration.}
    \label{fig:spike_distribution}
    \vspace{-4mm}
\end{figure*}

% \begin{figure}[ht!]
%     \centering    
    
%     \caption{Caption}
%     \label{fig:my_label}
% \end{figure}
%\vspace{-3mm}
\subsection{Price Spike Events}
For the statistical analysis of price spikes, the marginal cost of energy (MCE) is used as the price signal for both CAISO and ISO-NE, with 95$^{th}$ and 99$^{th}$ percentiles being the thresholds for the moderate- and high- spike points. 
% A \emph{price-spike point} is a point in time when MCE is higher than a threshold. 
Table~\ref{tab:spike_distribution} depicts the thresholds for moderate spikes (Q-95) and high spikes (Q-99), as computed from the CAISO and ISO-NE data.

\begin{table}[h!]
    \centering
    \vspace{-0.4cm}
    \caption{Spike Distribution}
    \label{tab:spike_distribution}
    \ra{1.3}
    \resizebox{\columnwidth}{!}{%
    \begin{tabular}{@{}l|cccccccccc@{}}
        \toprule
        ISO & \multicolumn{4}{c}{MCE ($\$$ per MW)} & \phantom{abc}& \multicolumn{5}{c}{Price Spike Events} \\
        \cmidrule{2-5} \cmidrule{7-11}
        & min & Q-95 & Q-99 & max && n & 2018 & 2019 & 2020 & 2021\\ 
        \midrule
        CAISO & -86.37 & 59.68 & 140.51 & 1152.71 && 1004 & 423 & 381 & 200 & - \\ 
        ISO-NE & -97.44 & 81.79 & 133.25 & 506.62 && 223 & - & - & 40 & 183 \\
        \bottomrule
    \end{tabular}
    }
\end{table}
%\vspace{-3mm}
High-spike points (moderate-spike points are considered normal in this study) are grouped together to identify a \emph{price-spike event}. For grouping, any two consecutive high-spike points which are at least 5-mins (one interval) apart are considered to belong to two separate price-spike events. Given this definition, 1004 price-spike events were identified for CAISO between 2018 and 2021, and 223 price-spike events were observed for ISO-NE in 2020 and 2021.

% \subsubsection*{Handling Missing Data}
% Segments with missing price values are dropped. After dropping those anomalous segments with missing price information, there exists 486 price spike segments.

% @Milan: Add two plots (distribution of spikes by time of day and season - CAISO and ISO-NE; distribution of spikes by spike length)

The following observations were made from the distributions of price spike events of CAISO and ISO-NE (see Fig.~\ref{fig:spike_distribution}). 
% Here, Winter implies Dec-Feb, spring implies Mar-May, summer implies Jun-Aug, and Fall implies Sep-Nov
\begin{itemize}
    \item While spring is the biggest price-spike season for CAISO, price-spikes are least probable in spring for ISO-NE. 
    \item Except for Winters, price spikes are highly probable during the evening time for CAISO across all the seasons. Though this statement is true for ISO-NE for the summer season, the spikes are almost equally probable at different times of the day for all other seasons.
    \item For both CAISO and ISO-NE, midday is the least probable time period across all the seasons for a price-spike event to occur.
    \item Most spike events in CAISO exist only for one 5-min interval, however, most spike events in ISO-NE stay for three 5-min intervals.
    \item Spikes longer than an hour was noticed in both CAISO and ISO-NE, which often is an indication of an extreme event (e.g. wildfires, hurricanes). Such events are more common in ISO-NE (20\%) than in CAISO (3\%).
    % \item On any typical day, prices are particularly high during morning and evening times. In the morning and the evening times, high movement in renewables adds uncertainty into the system; thus, leading to high forecasting errors from day ahead predictions. 
    % \item In the winter season, when the hydro generation is less, there are fewer number of spikes. 
    % \item More positive price spikes were observed in the spring and summer seasons than the fall and winter, which coincided with the higher contribution from renewable sources as well as increased forecasted load. 
\end{itemize}

The statistical analysis of the distribution of price-spike events across CAISO and ISO-NE indicates that primary drivers of a price-spike event can vary significantly depending on the season, time of the day, and location of the ISO. 
% The proposed framework automatically identifies and reports primary factors driving the price spike-events. 
Next, we discuss model training followed by primary drivers as identified by the framework for both CAISO and ISO-NE.

% Filtering the columns:
% - Kept only one MCE column (PGAE_RTM_MCE) because others are repetitive, and all RTM (to capture congestion in real-time) and DAM (to capture extreme events) MCC columns. Only use mean for DAM MCC. 
% - Dropped LMP and MCL columns
% data on 12/31/2020 is missing (skip that day)

% \input{tables/caiso_results.tex}

\subsection{Model Validation}
The random forest classifier trained on 67\% of segments achieved an accuracy of 92\% and 82\% on the remaining 33\% of test data for CAISO and ISO-NE, respectively. The train set and the test set included 4\% and 1.5\% anomalous segments for CAISO and ISO-NE, respectively. Table~\ref{tab:lodes} depicts the number of derived (raw) features for both the ISOs. Raw features were selected based on data availability and filtered using statistical analysis of the data and inputs from the subject matter expert (SME). In the table, the \emph{Others} column includes features related to the demand, solar/wind curtailment, and transfers (imports and exports). 
\begin{table}[ht!]
    \centering
    \vspace{-0.4cm}
    \caption{Feature Count: Derived (Raw)}
    \label{tab:lodes}
    % \scriptsize
    \resizebox{\columnwidth}{!}{%
    %\begin{tabular}{|l|p{2cm}|p{2cm}|p{2cm}|}
    \begin{tabular}{@{}l|cccccc@{}}
    \toprule
    Parameter & Prices && \multirow{2}{*}{\shortstack{Forecasting \\ Error}} & Generation & \multirow{2}{*}{\shortstack{Regulation \\ Prices}} & Others \\ [0.5ex]
    \cmidrule{2-3}
    & Congestion & Day-Ahead & &&& \\
    \midrule
    CAISO: 105(30) & 16(4) & 5(5) & 24(6) &  20(5) & 24(6) & 16(4) \\ \midrule
    ISO-NE: 170(56) & 36(9) & 17(17) & 9(3) & 40(10) & 56(14) & 12(3) \\
    \bottomrule
    \end{tabular}
    }
\end{table}

The false positive rate of the trained model is 13\% (769/5919) for CAISO and 1\% (42/4255) for ISO-NE. However, further analysis indicates that those segments are correlated with moderate price-spike events. In those events, CAISO and ISO-NE had mean energy costs of $\$51$/MWh and $\$91$/MWh, respectively, close to our observations in Table~\ref{tab:spike_distribution}.

\subsection{Primary Drivers}
\begin{figure}[h!]
    \centering
    \includegraphics[width=0.85\columnwidth]{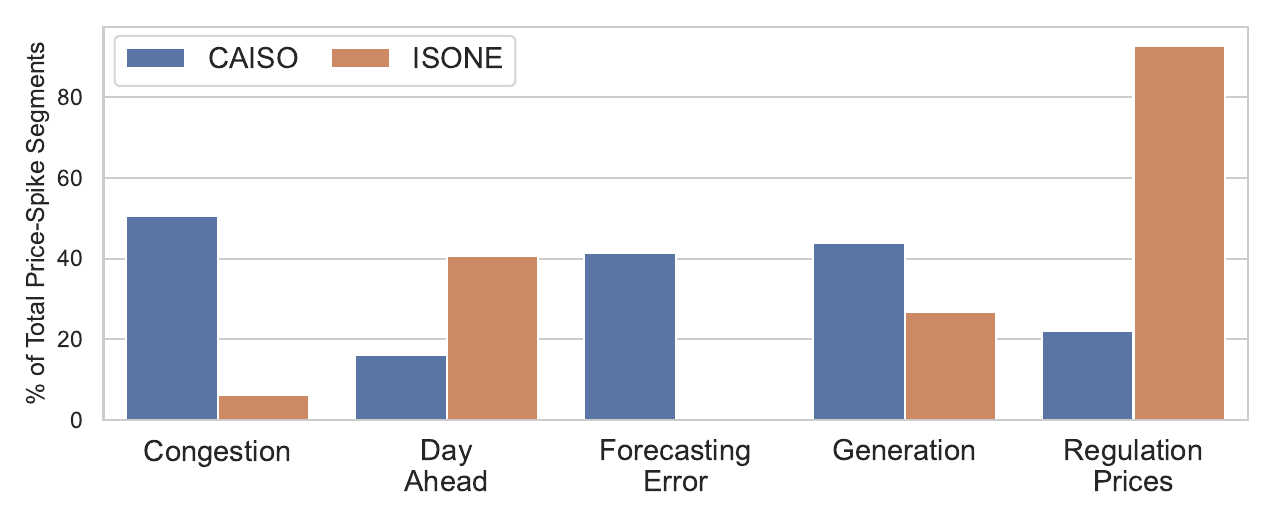}
    \vspace{-0.4cm}
    \caption{Key drivers behind price-spike segments}
    \label{fig:key_drivers}
    \vspace{-0.5cm}
\end{figure}

% Overall, the analysis indicates that depending on season and time of the day, different factors can be attributed to different instances of price spikes. Statistical analysis helped us in identifying key features that are highly correlated with the price spike intervals. While it was good for preliminary hypothesis testing, statistical analysis of significantly large number of raw and derived features could be a tedious and time-consuming task. 
Analysis of the top five features (from the red category) sorted by their contributions towards the predicted value indicates that the six key drivers (from Fig.~\ref{fig:key_drivers}) are highly correlated with the price-spike segments.

\begin{figure*}[ht!]
    \centering
    \includegraphics[width=0.8\textwidth]{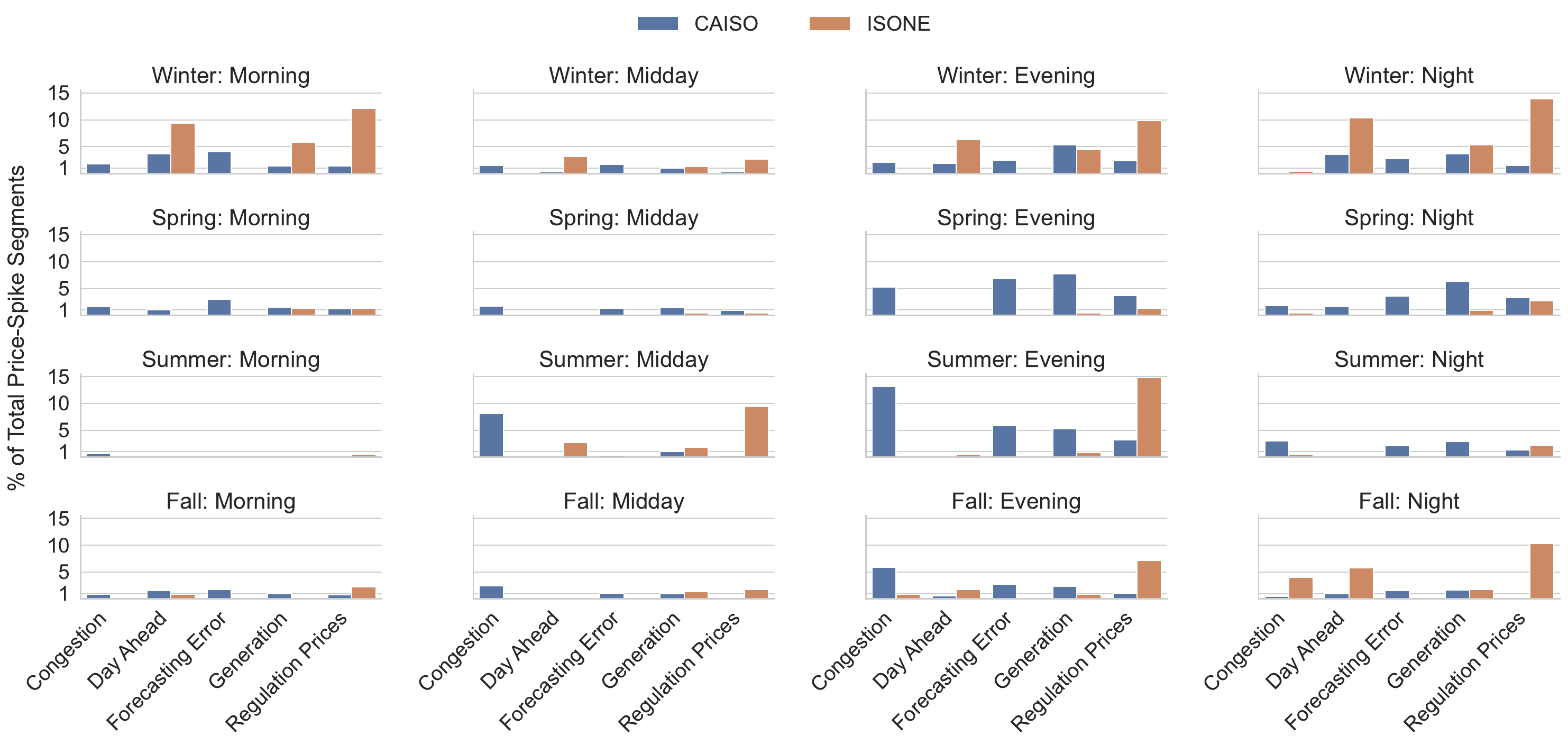}
    \vspace{-0.3cm}
    \caption{Breakdown of key drivers by Season and Daytime}
    \label{fig:drivers_by_season_daytime}
    \vspace{-3mm}
\end{figure*}
\begin{figure*}[ht!]
    \centering
    \includegraphics[width=0.95\textwidth]{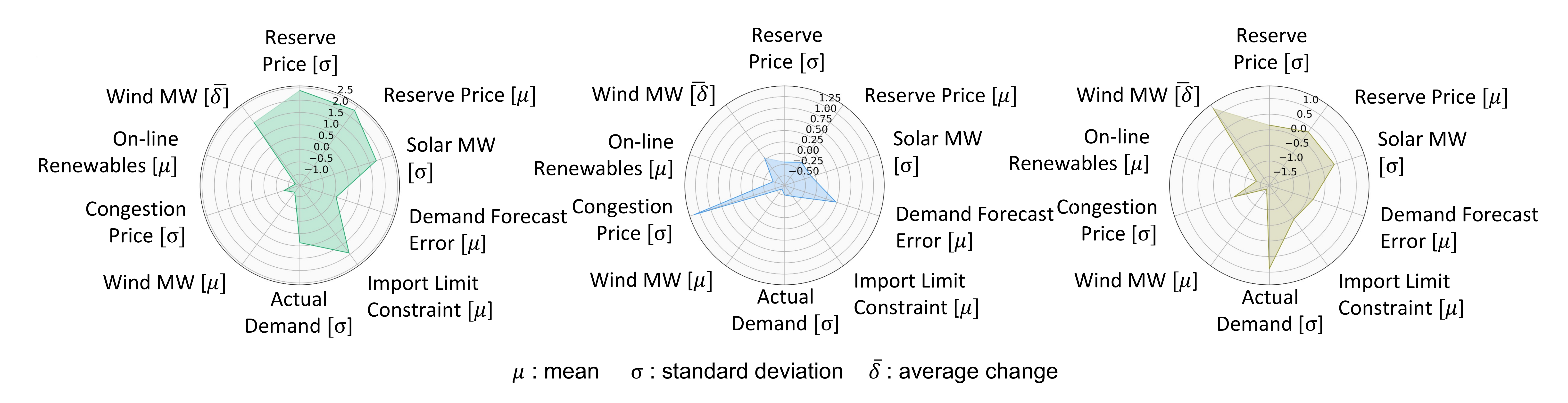}
    \vspace{-0.3cm}
    \caption{System states as identified by the K-Means for ISO-NE. [left] High Reserve Prices [middle] Congestion [right] Volatile Demand + High Renewables}
    \label{fig:clustering}
    \vspace{-0.4cm}
\end{figure*}
%\vspace{-3mm}
\subsubsection{Congestion (MCC$\ne$0)} Congestion captures average congestion cost and its movement during a segment in a real-time market. While it is a key driver for CAISO (50\%), it contributes only to a small proportion of spikes in ISO-NE (6\%). In CAISO, typically, the congestion between the Southern California nodes (SCE and VEA) and Northern California (PGAE) is a key reason, especially during the evening time across all the seasons and sometimes in midday during summer, as shown in Fig.~\ref{fig:drivers_by_season_daytime}.
% In the previous two examples, congestion was noticed whenever there is an abundance of resources. It can either happen due to unexpected increase in generation or unexpected reduction in demand. However, congestion can also happen for numerous other reasons. To capture those events, congestion is also defined as a separate root cause. 

\subsubsection{High Day-Ahead Prices}
Increase in gas prices, and extreme conditions (like wildfire or blackouts) are some of the several reasons in why system operator can anticipate high prices in the day-ahead market. Price spikes in such markets are hard to control or avoid. Such segments are more prominent in ISO-NE (41\%) than in CAISO (16\%). Price spikes under this category are more frequent in winter (Fig.~\ref{fig:drivers_by_season_daytime}) for both the ISOs. 
% Therefore, such events are kept in a separate bin. 

\subsubsection{Forecasting Error}
Market regulators forecast generation from renewables (mainly Solar and Wind) and demand in the day-ahead market. Though markets are often robust enough to handle forecasting errors, sometimes an unexpected generation or abrupt movement in resources can lead to price-spike. While this is a key driver for CAISO - 42\%, it is non-existent for ISO-NE. Further analysis of those events indicates that 90\% of price spikes under this category for CAISO can be attributed to forecasting errors in solar and wind projections. These spikes mainly exist during the evening times, as shown in Fig.~\ref{fig:drivers_by_season_daytime}, when solar ramps down. 
% ISO-NE dataset only contains wind forecast error, which could be a potential reason of low contribution of forecasting error in price-spikes for ISO-NE.

% Fig. 8 shows the distribution of renewables (mainly solar and wind), the day ahead forecast, and error with respect to the day-ahead forecast at different hours of the day (the x-axis) for all three years. The negative sign in the forecasting error indicates that the generation in real time was less than the forecasted value and vice-a-versa for the positive sign. During the morning time, when renewables ramp up, the solar and the wind generation remain low during the price spike intervals. Though the forecast is low too during those intervals, the gaps in the distribution of the forecasting error during spike v/s non-spike interval show that the system fails to anticipate how much lower.

\subsubsection{Movement in Generation}
Typically, movement in renewable sources (specifically solar, wind, and hydro) lead to a price-spike event, and movement in conventional resources (except Nuclear) respond to the deficit created by the movement of renewables. Fig.~\ref{fig:key_drivers} indicates that movement in the generation is a key driver in 44\% and 27\% price-spike segments for both CAISO and ISO-NE, respectively. While spikes in CAISO are more correlated with movement in renewables (Hydro-75\%, Solar, and Wind-25\%), spikes in ISO-NE can be attributed to movement in Oil-based generation (95\%). This further validates the limited role of renewables in ISO-NE, as also noticed previously in \emph{Forecasting Error}.
% On the other side, evening spikes were noticed to be more correlated to the change in renewables during consecutive time intervals. Fig. 10 shows the distribution of maximum change in renewables generation, the day ahead forecast, and the forecasting error during spike/non-spike intervals. Renewables ramp down rapidly during the evening and rapid change in renewable forecasting error can stress up the system which leads to ramping up of thermal resources.  

\subsubsection{Regulation Prices}
Regulation/clearing prices were a key driver in 93\% of the total price-spike segments in ISO-NE, with the primary reason being the movement in renewable capacity, present in 85\% of the total spike segments of this category. Fig.~\ref{fig:key_drivers} indicates that these spike segments occurred mainly in Winter (Morning, Evening, and Night), Summer (Midday and Evening), and Fall (Evening and Night). For CAISO, only 22\% of all spike segments are driven by regulation prices with the key reason being the volatility of Regulation Up prices (99\%). 
% Regulation prices is another key factor that is highly correlated with the price spike intervals. Fig. 12 shows the distribution of regulation prices during spike and non-spike intervals starting from 2018 (top) till 2020 (bottom). The outlier instances in the plot show intervals when the regulation prices were high. Compared to other factors, such instances are rare, but more evident during the evening time.

\subsection{Capturing the System State}
% A key take-away of the analysis is the fact that often multiple factors weigh-in during a price spike event and it is not feasible to single-out one key driver. Therefore, an unsupervisted clustering alogrithm (K-Means) is used to define system state based on SHAP values and spider charts to visualize those clusters.  
The framework offers an additional feature of data-driven categorization of potential system states using K-Means clustering and its visualization using radar plots. Through Elbow method~\cite{syakur2018integration} we identified 8 clusters for both the ISOs. Fig.~\ref{fig:clustering} depicts three such clusters from ISO-NE where each cluster is characterizing a specific system state: high reserve prices (left), congestion (middle), and volatile demand + high renewables (right).
The left radar chart shows that key variables like the mean value $\mu$ and the standard deviation $\sigma$ of the reserve prices are high during high renewable volatility (wind MW average change $\bar\delta$ and solar MW $\sigma$) generation period, leading to the additional requirement of regulation reserves to be procured and leading increase in energy prices. The middle radar chart represents a scenario where the congestion prices $\sigma$ are high, and demand forecast error $\mu$ are high leading to high prices in the market. The right radar chart shows how volatility in demand $\sigma$, wind MW generation $\bar\delta$, can lead to increased electricity market prices as expensive resources are being dispatched to cover for the uncertainty in load and renewable generation. Clusters, in association with key drivers, are crucial for the users to make informed decisions. 
\section{Concluding Remarks}
% In this paper, we 
% can be extended other kinds of price signatures, 
The primary contribution of this paper is an ML framework to automatically identify and report key factors driving the price spike events in the electricity markets. The framework was demonstrated on CAISO and ISO-NE and the analysis indicates that while congestion and renewable movement are key drivers for price-spike in CAISO, regulation prices, and day-ahead markets drive price spikes in ISO-NE. A high correlation of day-ahead prices with price-spike segments also explains longer spikes in ISO-NE, compared to CAISO. 
% Clustering and visualization through radar plots complement the analysis by helping the user understand the system state. 
Our analysis indicates that insights from ISO about a certain price pattern cannot be generalized to other ISOs. Key drivers behind certain price patterns can vary depending upon the ISO location, local weather conditions, time of the day, and several other reasons. In this context, the insights generated from such an automated analytical framework about the key drivers driving the price spikes can be used by market and system operators to analyze market conditions in near real-time. As of now, market operators analyze market data after the fact and perform price corrections and adjustments during the settlement process, which in real time is non-trivial for complex market data. The market operators can use the proposed framework to even cluster the market run results and validate it in near real-time. Furthermore, the automated labeling of price-spike events and assignment of specific reasons like congestion, renewable volatility, forecast errors, etc. as primary drivers can be utilized cyber-security experts to spot market attacks. Akin to that, market designers and policy analysts can use these insights to comprehend the market mechanisms for price formation to improve the forecast of electricity prices, bring transparency in the market operations, and design appropriate market-based interventions to mitigate such scenarios. 
% Data and code will be open-sourced as part of this study. 
In future, we intend to extend this analysis to other ISOs and also incorporate other price patterns, including price volatility. 

%\vspace{-3mm}
%\section*{Acknowledgment}

%The preferred spelling of the word ``acknowledgment'' in America is without 
%an ``e'' after the ``g''. Avoid the stilted expression ``one of us (R. B. 
%G.) thanks $\ldots$''. Instead, try ``R. B. G. thanks$\ldots$''. Put sponsor 
%acknowledgments in the unnumbered footnote on the first page.

%\section*{References}
\bibliographystyle{IEEEtran}
%\vspace{-2mm}
\bibliography{main.bib}

\end{document}